\documentclass[a4paper]{article}

\usepackage[english]{babel}
\usepackage[utf8]{inputenc}
\usepackage{amsmath}
\usepackage{graphicx}
\usepackage[colorinlistoftodos]{todonotes}

\usepackage[font=small,labelfont=bf]{caption}
\usepackage{abstract}
\usepackage{authblk}
\usepackage{booktabs}
\usepackage{lineno}
\newcommand*\samethanks[1][\value{footnote}]{\footnotemark[#1]}
\DeclareMathOperator{\DR}{DR}
\DeclareMathOperator{\PSNR}{PSNR}
\DeclareMathOperator{\SSIM}{SSIM}
\usepackage{todonotes}
\title{Convolutional Sketch Inversion} 

\author{Yağmur Güçlütürk\thanks{Y. Güçlütürk and U. Güçlü contributed equally to this work.}}
\author{Umut Güçlü\samethanks}
\author{Rob van Lier}
\author{Marcel A. J. van Gerven}
\affil{Radboud University, Donders Institute for Brain, Cognition and Behaviour, Nijmegen, the Netherlands}

\date{}

\begin{document}
\maketitle

\begin{figure}[h]
\centering
\includegraphics[width=\textwidth]{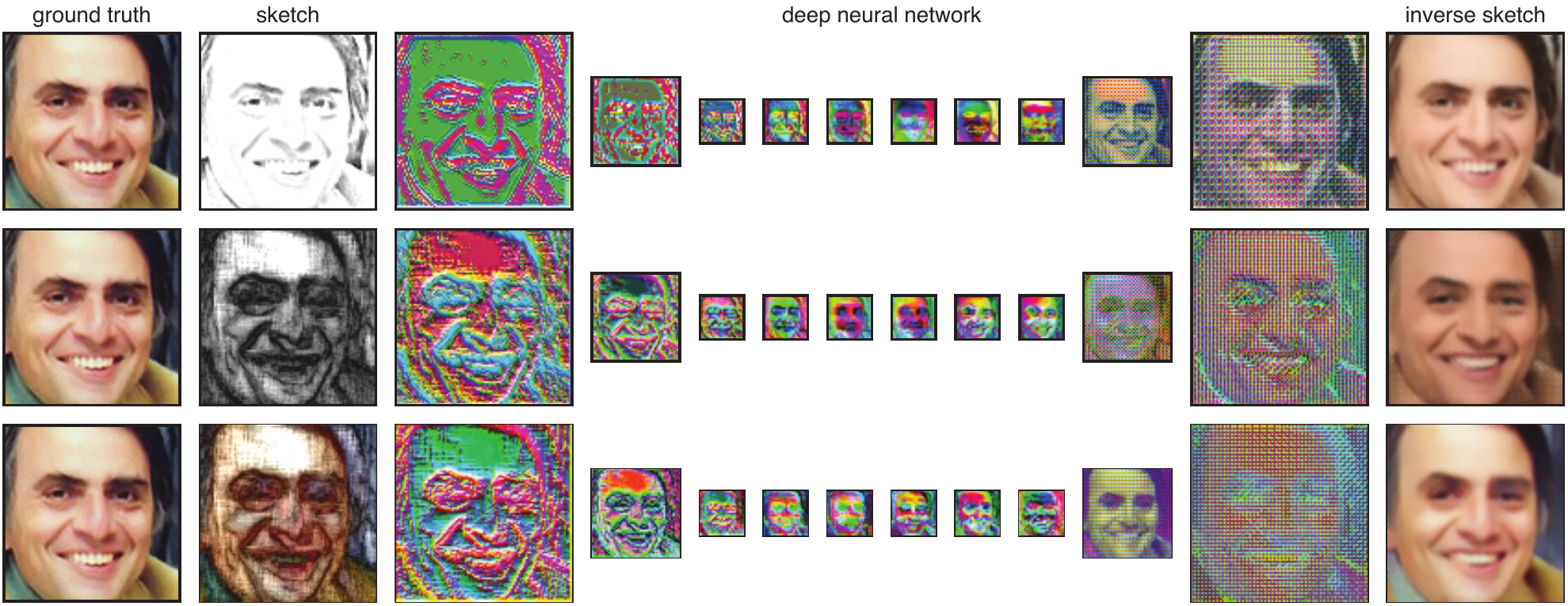}
\caption{Example results of our convolutional sketch inversion models. Our models invert face sketches to synthesize photorealistic face images. Each row shows the sketch inversion / photo synthesis pipeline that transforms a different sketch of the same face to a different image of the same face via a different deep neural network. Each deep neural network layer is represented by the top three principal components of its feature maps.}
\label{figure_1}
\end{figure}

\renewcommand{\abstractname}{}
\renewcommand{\absnamepos}{empty}
\begin{abstract}
\noindent \textbf{Abstract.} In this paper, we use deep neural networks for inverting face sketches to synthesize photorealistic face images. We first construct a semi-simulated dataset containing a very large number of computer-generated face sketches with different styles and corresponding face images by expanding existing unconstrained face data sets. We then train models achieving state-of-the-art results on both computer-generated sketches and hand-drawn sketches by leveraging recent advances in deep learning such as batch normalization, deep residual learning, perceptual losses and stochastic optimization in combination with our new dataset. We finally demonstrate potential applications of our models in fine arts and forensic arts. In contrast to existing patch-based approaches, our deep-neural-network-based approach can be used for synthesizing photorealistic face images by inverting face sketches in the wild.
\end{abstract}

\section{Introduction}

Portrait and self-portrait sketches have an important role in art. From an art historical perspective, self-portraits serve as historical records of what the artists looked like. From the perspective of an artist, self-portraits can be seen as a way to practice and improve one's skills without the need for a model to pose. Portraits of others further serve as memorabilia and a record of the person in the portrait. Artists most often are able to easily capture recognizable features of a person in their sketches. Therefore, hand-drawn sketches of people have further applications in law enforcement. Sketches of suspects drawn based on eye-witness accounts are used to identify suspects, either in person or from catalogues of mugshots.

Prior work related to face sketches in computer vision has been mostly limited to synthesis of highly controlled (i.e. having neutral expression, frontal pose, with normal lighting and without any occlusions) sketches from photographs~\cite{Tang2003,Liu2005,Gao2012,Wang2013b,Zhang2015} (sketch synthesis) and photographs from sketches~\cite{Liu2007,Xiao2009,Wang2009,Gao2012,Wang2013b} (sketch inversion). Sketch inversion studies with controlled inputs utilized patch-based approaches and used Bayesian tensor inference~\cite{Liu2007}, an embedded hidden Markov model~\cite{Xiao2009}, a multiscale Markov random field model~\cite{Wang2009}, sparse representations~\cite{Gao2012} and transductive learning with a probabilistic graph model~\cite{Wang2013b}. 

Few studies developed methods of sketch synthesis to handle more variation in one or more variables at a time, such as lighting~\cite{Li2006}, and lighting and pose~\cite{Zhang2010}. In a recent study, Zhang et al.~\cite{Zhang2016} showed that sketch synthesis by transferring the style of a single sketch could be used also in uncontrolled conditions. In~\cite{Zhang2016}, first an initial sketch by a sparse representation-based greedy search strategy was estimated, then candidate patches were selected from a template style sketch and the estimated initial sketch. Finally, the candidate patches were refined by a multi-feature-based optimization model and the patches were assembled to produce the final synthesized sketch.  

Recently, the use of deep convolutional neural networks (DNNs) in image transformation tasks, in which one type of image is transformed into another, has gained tremendous traction. In the context of sketch analysis, DNNs were used to tackle the problems of sketch synthesis and sketch simplification. For example,~\cite{Zhang2015} has used a DNN to convert photographs to sketches. They developed a DNN with six convolutional layers and a discriminative regularization term for enhancing the discriminability of the generated sketch against other sketches. Furthermore,~\cite{SimoSerra2016} has used a DNN to simplify rough sketches. They have shown that users prefer sketches simplified by the DNN more than they do those by other applications 97\% of the time.

Some other notable image transformation problems include colorization, style transfer and super-resolution. In colorization, the task is to transform a grayscale image to a color image that accurately captures the color information. In style transfer, the task is to transform one image to another image that captures the style of a third image. In super-resolution, the task is to transform a low-resolution image to a high-resolution image with maximum quality. DNNs have been used to tackle all of these problems with state-of-the art results~\cite{Cheng2015,Iizuka2016,Gatys2015,Dong2014,Dong2016,Johnson2016}.

However, a challenging task that remains is photorealistic face image synthesis from face sketches in uncontrolled conditions. That is, at present, there exist no sketch inversion models that are able to perform in realistic conditions. These conditions are characterized by changes in expression, pose, lighting condition and image quality, as well as the presence of varying amounts of background clutter and occlusions. 

Here, we use DNNs to tackle the problem of inverting face sketches to synthesize photorealistic face images from different sketch styles in uncontrolled conditions. We developed three different models to handle three different types of sketch styles by training DNNs on datasets that we constructed by extending a well-known large-scale face dataset, obtained in uncontrolled conditions~\cite{Liu2015}. We test the models on another similar large-scale dataset~\cite{LearnedMiller2016}, a hand-drawn sketch database~\cite{Wang2009} as well as on self-portrait sketches of famous Dutch artists. We show that our approach, which we refer to as {\em Convolutional Sketch Inversion} (CSI) can be used to achieve state-of-the-art results and discuss possible applications in fine arts, art history and forensics.

\section{Semi-simulated datasets}

For training and testing our CSI model, we made use of the following datasets:
\begin{itemize}
\item \textit{Large-scale CelebFaces Attributes (CelebA) dataset}~\cite{Liu2015}.
The CelebA dataset contains 202,599 celebrity face images and 10,177 identities. The images were obtained from the internet and vary extensively in terms of pose, expression, lighting, image quality, background clutter and occlusion. Each image in the dataset has five landmark positions and 40 attributes. These images were used for training the networks.

\item \textit{Labeled Faces in the Wild (LFW) dataset}~\cite{LearnedMiller2016}.
The LFW dataset contains 13,233 face images and 5749 identities. Similar to the CelebA dataset, images were obtained from the internet and vary extensively in terms of pose, expression, lighting, image quality, background clutter and occlusion. A subset of these images (11,990) were used for testing the networks.

\item \textit{CUHK Face Sketch (CUFS) database}~\cite{Wang2009}.
The CUFS database contains photographs and their corresponding hand-drawn sketches of 606 individuals. The dataset was formed by combining face photographs from three other databases and producing hand-drawn sketches of these photographs. Concretely, it consists of 188 face photographs  from the Chinese University of Hong Kong (CUHK) student database~\cite{Wang2009} and their corresponding sketches, 123 face photographs from the AR Face Database~\cite{Martinez1998} and their corresponding sketches, and 295 face photographs from the XM2VTS database~\cite{Messer1999} and their corresponding sketches. Only 18 of the sketches (six from each sub-database) were used in the current study. These images were used for testing the networks.

\item \textit{Sketches of famous Dutch artists}.
We also used the following sketches: i) Self-Portrait with Beret, Wide-Eyed by Rembrandt, 1630, etching, ii) Two Self-portraits and Several Details by Vincent van Gogh, 1886, pencil on paper and iii) Self-Portrait by M.C. Escher, 1929, lithograph on gray paper. These images were used for testing the networks. 
\end{itemize}

\subsection{Preprocessing}

Similar to~\cite{Cowen2014}, each image was cropped and resized to 96 pixels $\times$ 96 pixels such that:
\begin{itemize}
\item The distance between the top of the image and the vertical center of the eyes was 38 pixels.
\item The distance between the vertical center of the eyes and the vertical center of the mouth was 32 pixels.
\item The distance between the vertical center of the mouth and the bottom of the image was 26 pixels.
\item The horizontal center of the eyes and the mouth was at the horizontal center of the image.
\end{itemize}

\subsection{Sketching}

Each image in the CelebA and LFW datasets was automatically transformed to a line sketch, a grayscale sketch and a color sketch. Sketches in the CUFS database and those by the famous Dutch artists were further transformed to line sketches by using the same procedure.

Color and grayscale sketch types are produced by the same stylization algorithm~\cite{Gastal2011}. To obtain the sketch images, the input image is first filtered by an edge-aware filter. This filtered image is then blended with the magnitude of the gradient of the filtered image. Then, each pixel is scaled by a normalization factor resulting in the final sketch-like image.

Line sketches which resemble pencil sketches were generated based on~\cite{Beyeler2015}. Line sketch conversion works by first converting the color image to grayscale. This is followed by inverting the grayscale image to obtain a negative image. Next, a Gaussian blur is applied. Finally, using color dodge, the resulting image is blended with the grayscale version of the original image.

It should be noted that synthesizing face images from color or grayscale sketches is a more difficult problem than doing so from line sketches since many details of the faces are preserved by line sketches while they are lost for other sketch types.

\section{Models}

We developed one DNN for each of the three sketch styles based on the style transfer architecture in~\cite{Johnson2016}. Each of the three DNNs was based on the same architecture except for the first layer where the number of input channels were either one or three depending on the number of color channels of the sketches. The architecture comprised four convolutional layers, five residual blocks~\cite{He2015}, two deconvolutional layers and another convolutional layer. Each of the five residual blocks comprised two convolutional layers. All of the layers except for the last layer were followed by batch normalization~\cite{Ioffe2015} and rectified linear units. The last layer was followed by batch normalization and hyperbolic tangent units. All models were implemented in the Chainer framework~\cite{Tokui2015}. Table~\ref{table_1} shows the details of the architecture.

\begin{table}[]
\centering
\resizebox{\textwidth}{!}{%
\begin{tabular}{@{}lllllllll@{}}
\toprule
Layer & Type & in\_channels & out\_channels & ksize & stride & pad & normalization & activation \\ \midrule
1 & con. & 1 or 3 & 32 & 9 & 1 & 4 & BN & ReLU \\
2 & con. & 32 & 64 & 3 & 2 & 1 & BN & ReLU \\
3 & con. & 64 & 128 & 3 & 2 & 1 & BN & ReLU \\
4 & res. & 128/128 & 128/128 & 3/3 & 1/1 & 1/1 & BN/BN & ReLU \\
5 & res. & 128/128 & 128/128 & 3/3 & 1/1 & 1/1 & BN/BN & ReLU/+x \\
6 & res. & 128/128 & 128/128 & 3/3 & 1/1 & 1/1 & BN/BN & ReLU/+x \\
7 & res. & 128/128 & 128/128 & 3/3 & 1/1 & 1/1 & BN/BN & ReLU/+x \\
8 & res. & 128/128 & 128/128 & 3/3 & 1/1 & 1/1 & BN/BN & ReLU/+x \\
9 & dec. & 128 & 64 & 3 & 2 & 1 & BN & ReLU \\
10 & dec. & 64 & 32 & 3 & 2 & 1 & BN & ReLU \\
11 & con. & 32 & 3 & 9 & 1 & 4 & BN & tanh \\ \bottomrule
\end{tabular}%
}
\caption{Deep neural network architectures. BN; batch normalization with decay = 0.9, $\epsilon = 1e-5$, ReLU; rectified linear unit, con.; convolution, dec.; deconvolution, res.; residual block, tanh; hyperbolic tangent unit. Outputs of the hyperbolic tangent units are scaled to \lbrack0, 255\rbrack. x/y indicates the parameters of the first and second layers of a residual block. +x indicates that the input and output of a block are summed and no activation function is used.}
\label{table_1}
\end{table}

\subsection{Estimation}

For model optimization we used Adam~\cite{Kingma2014} with parameters $\alpha = 0.001$, $\beta_1 = 0.9$, $\beta_2 = 0.999$, $\epsilon = 10^{-8}$ and mini-batch size = 4. We trained the models by iteratively minimizing the loss function for 200,000 iterations. The loss function comprised three components. The first component is the standard Euclidean loss for the targets and the predictions (pixel loss; $\ell_p$). The second component is the Euclidean loss for the feature-transformed targets and the feature-transformed predictions (feature loss)~\cite{Johnson2016}:
\begin{equation}
\ell_{f} = \frac{1}{n}\sum_{i, j, k}\left(\phi\left(t\right)_{i, j, k} - \phi\left(y\right)_{i, j, k}\right) ^ 2
\end{equation}
where $n$ is the total number of features, $\phi(t)_{i, j, k}$ is a feature of the targets and $\phi(y)_{i, j, k}$ is a feature of the predictions. Similar to~\cite{Johnson2016}, we used the outputs of the fourth layer of a 16-layer DNN (relu\_2\_2 outputs of the VGG-16 pretrained model)~\cite{Simonyan2014} to feature transform the targets and the predictions. The third component is the total variation loss for the predictions:
\begin{equation}
\ell_{tv} = \sum_{i, j}\left(\left(y_{i + 1, j} - y_{i, j}\right) ^ 2 + \left(y_{i, j + 1} - y_{i, j}\right) ^ 2\right) ^ {0.5}
\end{equation}
where $y_{i, j}$ is a pixel of the predictions. A weighted combination of these components resulted in the following loss function:
\begin{equation}
\ell = \lambda_p \ell_p + \lambda_f \ell_f + \lambda_{tv} \ell_{tv}
\end{equation}
where we set $\lambda_p = \lambda_f = 1$ and $\lambda_{tv} = 0.00001$.

The use of the feature loss to train models for image transformation tasks was recently proposed by~\cite{Johnson2016}. In the context of super-resolution,~\cite{Johnson2016} found that replacing pixel loss with feature loss gives visually pleasing results at the expanse of image quality because of the artefacts introduced by the feature loss.

In the context of sketch inversion, our preliminary experiments showed that combining feature loss and pixel loss increases image quality while maintaining visual pleasantness. Furthermore, we observed that a small amount of total variation loss further removes the artefacts that are introduced by the feature loss. Therefore, we used the combination of the three losses in the final experiments. The quantitative results of the preliminary experiments in which the models were trained by using only the feature loss are provided in the Appendix.

\subsection{Validation}

\begin{figure}
\centering
\includegraphics[width=\textwidth]{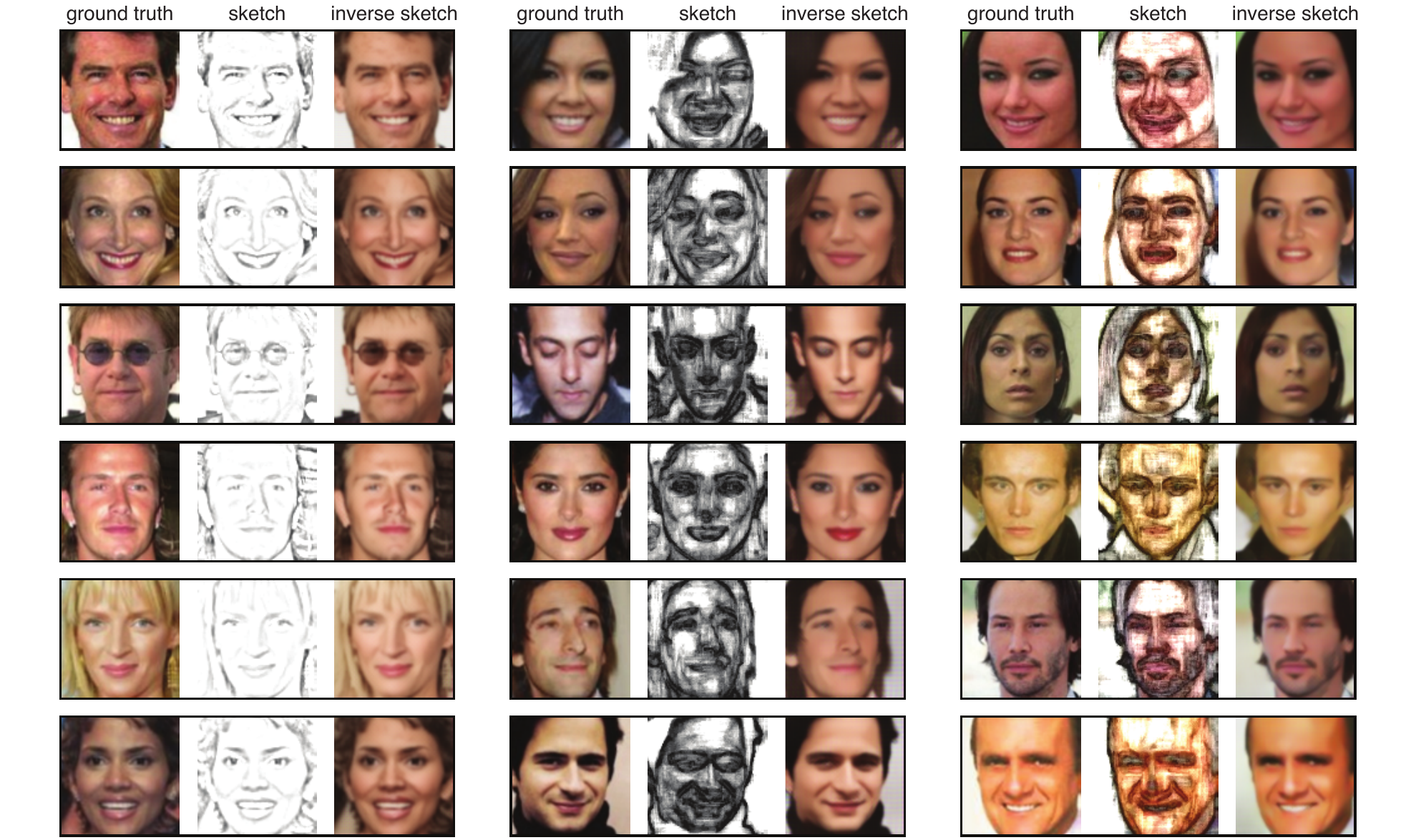}
\caption{Examples of the synthesized inverse sketches from the LFW dataset. Each distinct column shows examples from different sketch styles models, i.e. line sketch model (column 1), grayscale sketch model (column 2) and colour sketch model (column 3). First image in each column is the ground truth, the second image is the generated sketch and the third one is the synthesized inverse sketch.}
\label{figure_2}
\end{figure}

First, we qualitatively tested the models by visual inspection of the synthesized face images (Figure~\ref{figure_2}). Synthesized face images matched the ground truth photographs closely and persons in the images were easily recognizable in most cases. Among the three styles of sketch models, the line sketch model (Figure~\ref{figure_2}, first column) captured the highest level of detail in terms of the face structure, whereas the synthesized inverse sketches of the color sketch model (Figure~\ref{figure_2}, third column) had less structural detail but was able to better reproduce the color information in the ground truth images compared to the inverted sketches of the line sketch model. Sketches synthesized by the grayscale model (Figure~\ref{figure_2}, second column) were less detailed than those synthesized by the line sketch model. Furthermore, the color content was less accurate in sketches synthesized by the grayscale model than those synthesized by both the color sketch and the line sketch models. We found that the line model performed impressively in terms of matching the hair and skin color of the individuals even when the line sketches did not contain any color information. This may indicate that along with taking advantage of the luminance differences in the sketches to infer coloring, the model was able to learn color properties often associated with high-level face features of different ethnicities.

Then, we quantitatively tested the models by comparison of the peak signal to noise ratio (PSNR), structural similarity (SSIM) and standard Pearson product-moment correlation coefficient R of the synthesized face images~\cite{Wang2004} (Table~\ref{table_2}). PSNR measures the physical quality of an image. It is defined as the ratio between the peak power of the image and the power of the noise in the image (Euclidean distance between the image and the reference image):

\begin{equation}
\PSNR = \frac{1}{3} \sum_{k} 10 \log_{10}\frac{\max \DR ^ 2}{\frac{1}{m}\sum_{i, j}\left(t_{i, j, k} - y_{i, j, k}\right) ^ 2}
\end{equation}
where $\DR$ is the dynamic range, and $m$ is the total number of pixels in each of the three color channels. SSIM measures the perceptual quality of an image. It is defined as the multiplicative combination of the similarities between the image and the reference image in terms of contrast, luminance and structure: 

\begin{equation}
\SSIM = \frac{1}{3} \sum_{k} \frac{1}{m} \sum_{i, j} \frac{\left(2 \mu\left(t_{i, j, k}\right) \mu\left(y_{i, j, k}\right) + C_1\right) \left(2 \sigma\left(t_{i, j, k}, y_{i, j, k}\right) C_2\right)}{\left(\mu\left(t_{i, j, k}\right) ^ 2 \mu\left(y_{i, j, k}\right) ^ 2 + C_1\right) \left(2 \sigma\left(t_{i, j, k}\right) ^ 2 \sigma\left(y_{i, j, k}\right) ^ 2 C_2\right)}
\end{equation}
where $\mu\left(t_{i, j, k}\right)$, $\mu\left(y_{i, j, k}\right)$, $\sigma\left(t_{i, j, k}\right)$, $\sigma\left(y_{i, j, k}\right)$ and $\sigma\left(t_{i, j, k}, y_{i, j, k}\right)$ are means, standard deviations and cross-covariances of windows centered around $i$ and $j$. Furthermore, $C_1 = (0.01 \max \DR) ^ 2$ and $C_2 = (0.03 \max \DR) ^ 2$. Quality of a dataset is defined as the mean quality over the images in the dataset.

The inversion of the line sketches resulted in the highest quality face images for all three measures (20.12 for PSNR, 0.86 for SSIM and 0.93 for R). In contrast the inversion of the grayscale sketches resulted in the lowest quality face images for all measures (17.65 for PSNR, 0.65 for SSIM and 0.75 for R). This shows that both the physical and the perceptual quality of the inverted sketch images produced by the line sketch network was superior than those by the other sketch styles.

\begin{table}[]
\centering
\resizebox{\textwidth}{!}{%
\begin{tabular}{@{}lllllllllllll@{}}
\toprule
 &  &  &  & \multicolumn{1}{c}{\textit{PSNR}} &  &  &  & \multicolumn{1}{c}{\textit{SSIM}} &  &  &  & \multicolumn{1}{c}{\textit{R}} \\ \midrule
\textit{Line} &  &  &  & \textbf{20.1158 $\pm$ 0.0231} &  &  &  & \textbf{0.8583 $\pm$ 0.0003} &  &  &  & \textbf{0.9298 $\pm$ 0.0005} \\
\textit{Grayscale} &  &  &  & 17.6567 $\pm$ 0.0263 &  &  &  & 0.6529 $\pm$ 0.0008 &  &  &  & 0.7458 $\pm$ 0.0020 \\
Color &  &  &  & 19.2029 $\pm$ 0.0293 &  &  &  & 0.7154 $\pm$ 0.0008 &  &  &  & 0.8087 $\pm$ 0.0017 \\ \bottomrule
\end{tabular}
}
\caption{Comparison of physical (PSNR), perceptual (SSIM) and correlational (R) quality measures for the inverse sketches synthesized by the line, grayscale and color sketch-style models. $x \pm m$ shows the mean $\pm$ the bootstrap estimate of the standard error of the mean.}
\label{table_2}
\end{table}

Finally, we tested how well the line sketch inversion model can be transferred to the task of synthesizing face images from sketches that are hand-drawn and not generated using the same methods that were used to train the model. We considered only the line sketch model since the contents of the hand-drawn sketch database that we used~\cite{Wang2009} were most similar to the line sketches.

\begin{figure}
\centering
\includegraphics[width=\textwidth]{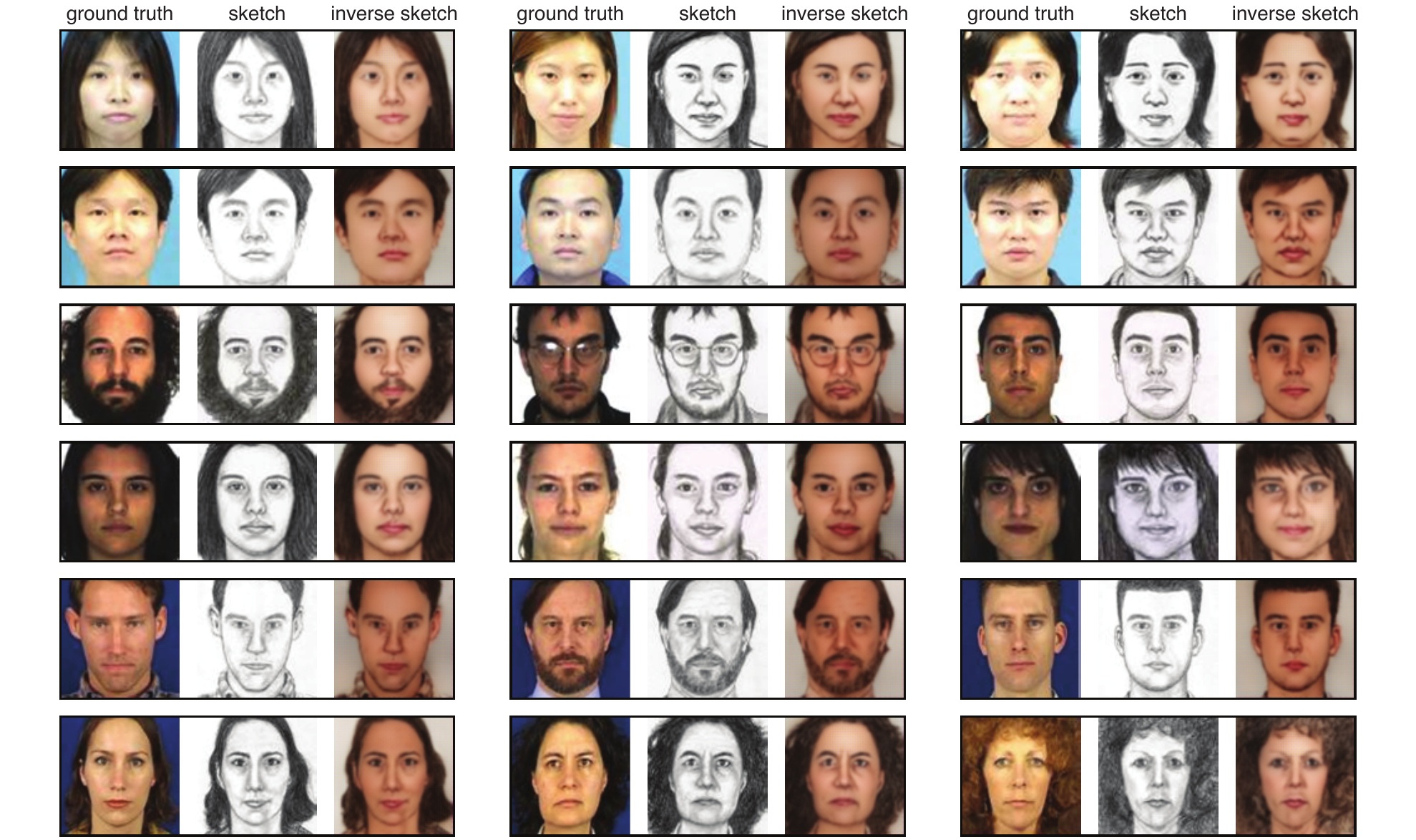}
\caption{Examples of the synthesized inverse sketches from the CUFS database. First image in each column is the ground truth, the second image is the sketch hand-drawn by an artist and the third one is the inverse sketch that was synthesized by the line sketch model.}
\label{figure_3}
\end{figure}

We found that the line sketch inversion model can solve this inductive transfer task almost as good as it can solve the task that it was trained on (Figure~\ref{figure_3}). Once again, the model synthesized photorealistic face images. While color was not always synthesized accurately, other elements such as form, shape, line, space and texture were often synthesized well. Furthermore hair texture and style, which posed a problem in most previous studies, was very well handled by our CSI model. We observed that the dark-edged pencil strokes in the hand-drawn sketches that were not accompanied by shading resulted in less realistic inversions (compare e.g nose areas of sketches in the first and second rows with those in the third row in Figure~\ref{figure_3}). This can be explained by the lack of such features in the training data of the line sketch model, and can be easily overcome by including training examples more closely resembling the drawing style of the sketch artists. 

\begin{table}[]
\centering
\resizebox{\textwidth}{!}{%
\begin{tabular}{@{}lllllllllllll@{}}
\toprule
 &  &  &  & \multicolumn{1}{c}{\textit{PSNR}} &  &  &  & \multicolumn{1}{c}{\textit{SSIM}} &  &  &  & \multicolumn{1}{c}{\textit{R}} \\ \midrule
\textit{CUHK (6)} &  &  &  & \textbf{15.0675 $\pm$ 0.3958} &  &  &  & 0.5658 $\pm$ 0.0099 &  &  &  & \textbf{0.8264 $\pm$ 0.0269} \\
\textit{AR (6)} &  &  &  & 13.8687 $\pm$ 0.7009 &  &  &  & \textbf{0.5684 $\pm$ 0.0277} &  &  &  & 0.7667 $\pm$ 0.0314 \\
XM2GTS (6) &  &  &  & 11.3293 $\pm$ 1.2156 &  &  &  & 0.4231 $\pm$ 0.0272 &  &  &  & 0.4138$\pm$ 0.1130 \\
\textit{All (18)} &  &  &  & 13.4218 $\pm$ 0.6123 &  &  &  & 0.5191 $\pm$ 0.0207 &  &  &  & 0.6690 $\pm$ 0.0591 \\ \bottomrule
\end{tabular}
}
\caption{Comparison of physical (PSNR), perceptual (SSIM) and correlational (R) quality measures for the inverse sketches synthesized from the sketches in the CUFS database and its sub-databases. $x \pm m$ shows the mean $\pm$ the bootstrap estimate of the standard error of the mean.}
\label{table_3}
\end{table}

\begin{figure}
\centering
\includegraphics[width=\textwidth]{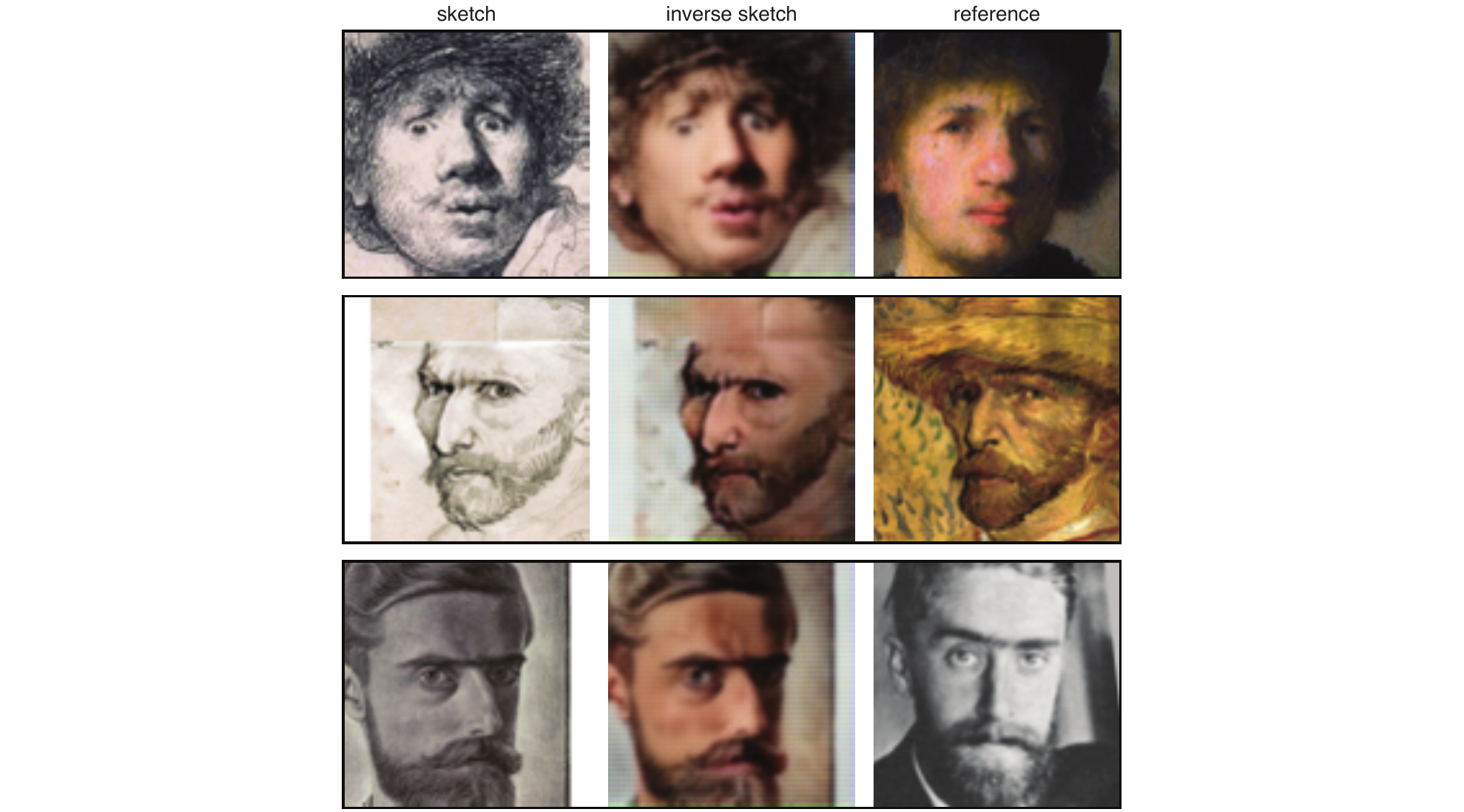}
\caption{Self-portrait sketches and synthesized inverse sketches along with a reference painting or photograph of famous Dutch artists: Rembrandt (top), Vincent van Gogh (middle) and M. C. Escher (bottom). Sketches: i) Self-Portrait with Beret, Wide-Eyed by Rembrandt, 1630, etching. ii) Two Self-portraits and Several Details by Vincent van Gogh, 1886, pencil on paper. iii) Self-Portrait by M.C. Escher, 1929, lithograph on gray paper. Reference paintings: i) Self-Portrait by Rembrandt, 1630, oil painting on copper. ii) Self-Portrait with Straw Hat by Vincent van Gogh, 1887, oil painting on canvas.}
\label{figure_4}
\end{figure}

For all the samples from the CUFS database, the PSNR, the SSIM index and the R of the synthesized face images were 13.42, 0.52, and 0.67, respectively (Table~\ref{table_3}). Among the three sub-databases of the CUFS database, the quality of the synthesized images from the CUHK dataset was the highest in terms of the PSNR (15.07) and R (0.83). While the PSNR and R values for the AR dataset was lower than those of the CUHK dataset, SSIM did not differ between the two datasets. The lowest quality inverted sketches were produced from the sample sketches of the XM2GTS database (with 13.42 for PSNR, 0.42 for SSIM and 0.41 for R).

\section{Applications}
\subsection{Fine arts}

In many cases self-portrait studies allow us a glimpse of what famous artists looked like through the artists' own perspective. Since there are no photographic records of many artists (in particular of those who lived before the 19th century during which the photography was invented and became widespread) self-portrait sketches and paintings are the only visual records that we have of many artists. Converting the sketches of the artists into photographs using a DNN that was trained on tens of thousands of face sketch-photograph pairs results in very interesting end-products.

\begin{figure}
\centering
\includegraphics[width=\textwidth]{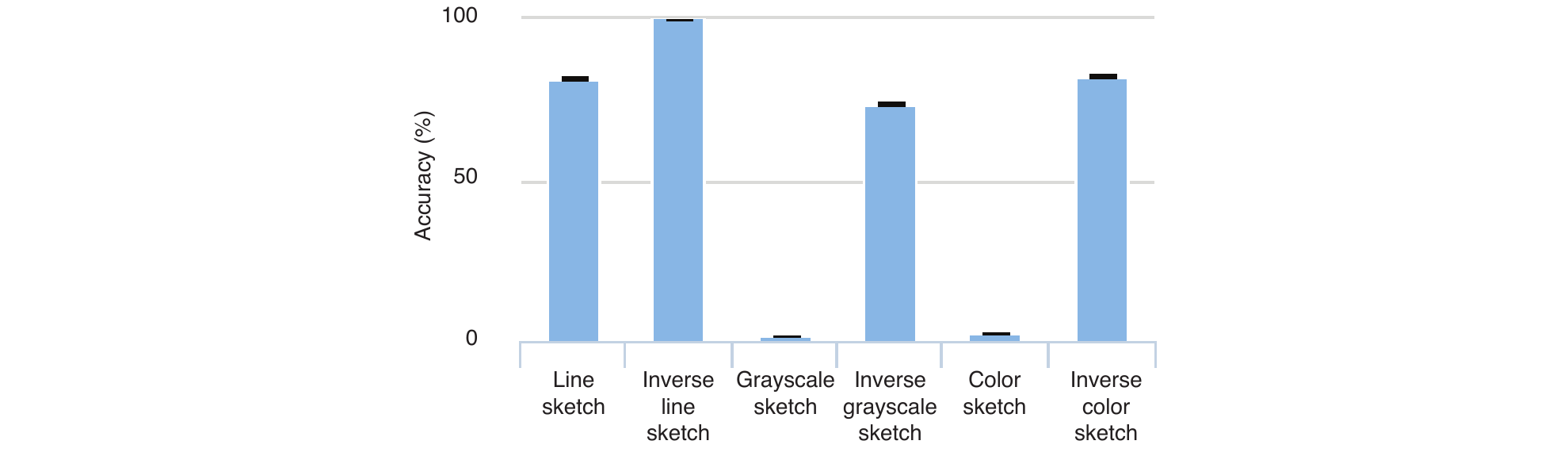}
\caption{Identification accuracies for line, grayscale and color sketches, and for inverse sketches synthesized by the corresponding models. Error bars show the bootstrap estimates of the standard errors.}
\label{figure_5}
\end{figure}

Here we used our DNN-based approach to synthesize photographs of famous Dutch artists Rembrandt, Vincent van Gogh and M. C. Escher from their self-portrait sketches\footnote{For simplicity, although different methods were used to produce these artworks, we refer to them as sketches.} (Figure~\ref{figure_4}). To the best of our knowledge, the synthesized photorealistic images of these artists are the first of their kind.

Our qualitative assesment revealed that, the inverted sketch of Rembrandt synthesized from his 1630 sketch indeed resembles himself in his paintings (particulary his self-portrait painting from 1630), and Escher's to his photographs. We found that the inverted sketch of van Gogh synthesized from his 1886 sketch was the most realistic synthesized photograph among those of the three artists, albeit not closely matching his self-portrait paintings of a distinct post-impressionist style.

Although we do not have a quantitative way to measure the accuracy of the results in this case, results demonstrate that the artistic style of the input sketches influence the quality of the produced photorealistic images. Generating new training sketch data to match more closely to the sketch style of a specific artist of interest (e.g. by using the method proposed by~\cite{Zhang2016}), and training the network with these sketches would overcome this limitation.

Sketching is one of the most important training methods that artist use to develop their skills. Converting sketches into photorealistic images would allow the artists in training to see and evaluate the accuracy of their sketches clearly and easily which can in turn become an efficient training tool. Furthermore, sketching is often much faster than producing a painting. When for example the sketch is based on imagination rather than a photograph, deep sketch inversion can provide a photorealistic guideline (or even an end-product, if digital art is being produced) and can speed up the production process of artists. Figure~\ref{figure_3}, which shows the inverted sketches by contemporary artists that produced the sketches in the CUFS database, further demonstrates this type of application. The current method can be developed into a smartphone/tablet or computer application for common use. 

\subsection{Forensic arts}

In cases where no other representation of a suspect exists, sketches drawn by forensic artists based on eye-witness accounts are frequently used by the law enforcement. However, direct use of sketches for automatically identifying suspects from databases containing photographs does not work well because these two face representations are too different to allow a direct comparison~\cite{Wang2013a}. Inverting a sketch to a photograph makes this task much easier by reducing the difference between these two alternative representations, enabling a direct automatized comparison~\cite{Wang2009}.

To evaluate the potential use of our system for forensic applications, we performed an identification analysis (Figure~\ref{figure_5}). In this analysis, we evaluated the accuracy of identifying a target face image in a very large set of candidate face images (LFW dataset containing over 11,000 images) from an (inverse) face sketch. The identification accuracies for the synthesized faces were always significantly higher than those for the corresponding sketched faces ($p << 0.05$, binomial test). While the identification accuracies for the color and grayscale sketches were very low (2.38\% and 1.42\%, respectively), those for the synthesized color and grayscale inverse sketches were relatively high (82.29\% and 73.81\%, respectively). On the other hand, identification accuracy of line sketches was already high, at 81.14\% before inversion. Synthesizing inverse sketches from line sketches raised the identification accuracy to an almost perfect level (99.79\%).

\section{Conclusions}

In this study we developed sketch datasets, complementing well known unconstrained benchmarking datasets~\cite{Liu2015, LearnedMiller2016}, developed DNN models that can synthesize face images from sketches with state-of-the-art performance and proposed applications of our CSI model in fine arts, art history and forensics. We foresee further computer vision applications of the developed methods for non-face images and various other sketch-like representations, as well as cognitive neuroscience applications for the study of cognitive phenomena such as perceptual filling in~\cite{Vergeer2015, Anstis2012} and the neural representation of complex stimuli~\cite{Guclu2015a, Guclu2015b}.

\clearpage

\bibliographystyle{ieeetr}
\bibliography{main}

\begin{thebibliography}{10}

\bibitem{Tang2003}
X.~Tang and X.~Wang, ``Face sketch synthesis and recognition,'' in {\em
  International Conference on Computer Vision}, Institute of Electrical {\&}
  Electronics Engineers ({IEEE}), 2003.

\bibitem{Liu2005}
Q.~Liu, X.~Tang, H.~Jin, H.~Lu, and S.~Ma, ``A nonlinear approach for face
  sketch synthesis and recognition,'' in {\em Conference on Computer Vision and
  Pattern Recognition}, Institute of Electrical {\&} Electronics Engineers
  ({IEEE}), 2005.

\bibitem{Gao2012}
X.~Gao, N.~Wang, D.~Tao, and X.~Li, ``Face sketch-photo synthesis and retrieval
  using sparse representation,'' {\em {IEEE} Transactions on Circuits and
  Systems for Video Technology}, vol.~22, pp.~1213--1226, aug 2012.

\bibitem{Wang2013b}
N.~Wang, D.~Tao, X.~Gao, X.~Li, and J.~Li, ``Transductive face sketch-photo
  synthesis,'' {\em {IEEE} Transactions on Neural Networks and Learning
  Systems}, vol.~24, pp.~1364--1376, sep 2013.

\bibitem{Zhang2015}
L.~Zhang, L.~Lin, X.~Wu, S.~Ding, and L.~Zhang, ``End-to-end photo-sketch
  generation via fully convolutional representation learning,'' in {\em
  International Conference on Multimedia Retrieval}, Association for Computing
  Machinery ({ACM}), 2015.

\bibitem{Liu2007}
W.~Liu, X.~Tang, and J.~Liu, ``Bayesian tensor inference for sketch-based
  facial photo hallucination,'' in {\em International Joint Conference on
  Artificial Intelligence}, 2007.

\bibitem{Xiao2009}
B.~Xiao, X.~Gao, D.~Tao, and X.~Li, ``A new approach for face recognition by
  sketches in photos,'' {\em Signal Processing}, vol.~89, pp.~1576--1588, aug
  2009.

\bibitem{Wang2009}
X.~Wang and X.~Tang, ``Face photo-sketch synthesis and recognition,'' {\em
  {IEEE} Transactions on Pattern Analysis and Machine Intelligence}, vol.~31,
  pp.~1955--1967, nov 2009.

\bibitem{Li2006}
Y.-h. Li, M.~Savvides, and V.~Bhagavatula, ``Illumination tolerant face
  recognition using a novel face from sketch synthesis approach and advanced
  correlation filters,'' in {\em International Conference on Acoustics, Speech,
  and Signal Processing}, Institute of Electrical {\&} Electronics Engineers
  ({IEEE}), 2006.

\bibitem{Zhang2010}
W.~Zhang, X.~Wang, and X.~Tang, ``Lighting and pose robust face sketch
  synthesis,'' in {\em European Conference on Computer Vision}, 2010.

\bibitem{Zhang2016}
S.~Zhang, X.~Gao, N.~Wang, and J.~Li, ``Robust face sketch style synthesis,''
  {\em {IEEE} Transactions on Image Processing}, vol.~25, pp.~220--232, jan
  2016.

\bibitem{SimoSerra2016}
E.~Simo-Serra, S.~Iizuka, K.~Sasaki, and H.~Ishikawa, ``Learning to simplify:
  Fully convolutional networks for rough sketch cleanup,'' {\em {ACM}
  Transactions on Graphics}, vol.~35, no.~4, 2016.

\bibitem{Cheng2015}
Z.~Cheng, Q.~Yang, and B.~Sheng, ``Deep colorization,'' in {\em International
  Conference on Computer Vision}, Institute of Electrical {\&} Electronics
  Engineers ({IEEE}), dec 2015.

\bibitem{Iizuka2016}
S.~Iizuka, E.~Simo-Serra, and H.~Ishikawa, ``Let there be color!: Joint
  end-to-end learning of global and local image priors for automatic image
  colorization with simultaneous classification,'' {\em {ACM} Transactions on
  Graphics}, vol.~35, no.~4, 2016.

\bibitem{Gatys2015}
L.~A. Gatys, A.~S. Ecker, and M.~Bethge, ``A neural algorithm of artistic
  style,'' {\em CoRR}, vol.~abs/1508.06576, 2015.

\bibitem{Dong2014}
K.~H. X.~T. Chao~Dong, Chen Change~Loy, ``Learning a deep convolutional network
  for image super-resolution,'' in {\em European Conference on Computer
  Vision}, 2014.

\bibitem{Dong2016}
C.~Dong, C.~C. Loy, K.~He, and X.~Tang, ``Image super-resolution using deep
  convolutional networks,'' {\em {IEEE} Transactions on Pattern Analysis and
  Machine Intelligence}, vol.~38, pp.~295--307, feb 2016.

\bibitem{Johnson2016}
J.~Johnson, A.~Alahi, and L.~Fei-Fei, ``Perceptual losses for real-time style
  transfer and super-resolution,'' {\em CoRR}, vol.~abs/1603.08155, 2016.

\bibitem{Liu2015}
L.~P. W.~X. Liu, Ziwei and X.~Tang, ``Deep learning face attributes in the
  wild,'' in {\em International Conference on Computer Vision}, 2015.

\bibitem{LearnedMiller2016}
E.~Learned-Miller, G.~B. Huang, A.~RoyChowdhury, H.~Li, and G.~Hua, ``Labeled
  faces in the wild: A survey,'' in {\em Advances in Face Detection and Facial
  Image Analysis}, pp.~189--248, Springer Science + Business Media, 2016.

\bibitem{Martinez1998}
A.~M. Martinez and R.~Benavente, ``The {AR}-face database.'' CVC Technical
  Report 24, 1998.

\bibitem{Messer1999}
K.~Messer, J.~Matas, J.~Kittler, and K.~Jonsson, ``Xm2vtsdb: The extended m2vts
  database,'' in {\em Audio and Video-based Biometric Person Authentication},
  1999.

\bibitem{Cowen2014}
A.~S. Cowen, M.~M. Chun, and B.~A. Kuhl, ``Neural portraits of perception:
  Reconstructing face images from evoked brain activity,'' {\em {NeuroImage}},
  vol.~94, pp.~12--22, jul 2014.

\bibitem{Gastal2011}
E.~S.~L. Gastal and M.~M. Oliveira, ``Domain transform for edge-aware image and
  video processing,'' {\em {ACM} Transactions on Graphics}, vol.~30, p.~1, jul
  2011.

\bibitem{Beyeler2015}
M.~Beyeler, {\em OpenCV with Python Blueprints}.
\newblock Birmingham, UK: Packt Publishing, 2015.

\bibitem{He2015}
K.~He, X.~Zhang, S.~Ren, and J.~Sun, ``Deep residual learning for image
  recognition,'' {\em CoRR}, vol.~abs/1512.03385, 2015.

\bibitem{Ioffe2015}
S.~Ioffe and C.~Szegedy, ``Batch normalization: Accelerating deep network
  training by reducing internal covariate shift,'' {\em CoRR},
  vol.~abs/1502.03167, 2015.

\bibitem{Tokui2015}
S.~Tokui, K.~Oono, S.~Hido, and J.~Clayton, ``Chainer: A next-generation open
  source framework for deep learning,'' in {\em Workshop on Machine Learning
  Systems at Neural Information Processing Systems}, 2015.

\bibitem{Kingma2014}
D.~Kingma and J.~Ba, ``Adam: A method for stochastic optimization,'' {\em
  CoRR}, vol.~abs/1412.6980, 2014.

\bibitem{Simonyan2014}
K.~Simonyan and A.~Zisserman, ``Very deep convolutional networks for
  large-scale image recognition,'' {\em CoRR}, vol.~abs/1409.1556, 2014.

\bibitem{Wang2004}
Z.~Wang, A.~Bovik, H.~Sheikh, and E.~Simoncelli, ``Image quality assessment:
  From error visibility to structural similarity,'' {\em {IEEE} Transactions on
  Image Processing}, vol.~13, pp.~600--612, apr 2004.

\bibitem{Wang2013a}
N.~Wang, D.~Tao, X.~Gao, X.~Li, and J.~Li, ``A comprehensive survey to face
  hallucination,'' {\em International Journal of Computer Vision}, vol.~106,
  pp.~9--30, aug 2013.

\bibitem{Vergeer2015}
M.~Vergeer, S.~Anstis, and R.~van Lier, ``Flexible color perception depending
  on the shape and positioning of achromatic contours,'' {\em Frontiers in
  Psychology}, vol.~6, may 2015.

\bibitem{Anstis2012}
S.~Anstis, M.~Vergeer, and R.~V. Lier, ``Looking at two paintings at once:
  Luminance edges can gate colors,'' {\em i-Perception}, vol.~3, no.~8,
  pp.~515--518, 2012.

\bibitem{Guclu2015a}
U.~G\"{u}{\c{c}}l\"{u} and M.~A.~J. van Gerven, ``Deep neural networks reveal a
  gradient in the complexity of neural representations across the ventral
  stream,'' {\em Journal of Neuroscience}, vol.~35, pp.~10005--10014, jul 2015.

\bibitem{Guclu2015b}
U.~G\"{u}{\c{c}}l\"{u} and M.~A.~J. van Gerven, ``Increasingly complex
  representations of natural movies across the dorsal stream are shared between
  subjects,'' {\em {NeuroImage}}, dec 2015. In Press.

\end{thebibliography}

\clearpage

\section{Appendix}

\begin{table}[h]
\centering
\resizebox{\textwidth}{!}{%
\begin{tabular}{@{}lllllllllllll@{}}
\toprule
 &  &  &  & \multicolumn{1}{c}{\textit{PSNR}} &  &  &  & \multicolumn{1}{c}{\textit{SSIM}} &  &  &  & \multicolumn{1}{c}{\textit{R}} \\ \midrule
\textit{Line} &  &  &  & 14.8956 $\pm$ 0.0207 &  &  &  & 0.5931 $\pm$ 0.0006 &  &  &  & 0.6023 $\pm$ 0.0017 \\
\textit{Grayscale} &  &  &  & 17.1654 $\pm$ 0.0277 &  &  &  & 0.6301 $\pm$ 0.0008 &  &  &  & 0.7175 $\pm$ 0.0022 \\
Color &  &  &  & \textbf{18.9884 $\pm$ 0.0296} &  &  &  & \textbf{0.7072 $\pm$ 0.0008} &  &  &  & \textbf{0.7976 $\pm$ 0.0019} \\ \bottomrule
\end{tabular}
}
\caption{Comparison of physical (PSNR), perceptual (SSIM) and correlational (R) quality measures for the inverse sketches synthesized by the line, grayscale and color sketch-style models trained using feature loss alone. $x \pm m$ shows the mean $\pm$ the bootstrap estimate of the standard error of the mean.}
\label{table_4}
\end{table}

\begin{table}[h]
\centering
\resizebox{\textwidth}{!}{%
\begin{tabular}{@{}lllllllllllll@{}}
\toprule
 &  &  &  & \multicolumn{1}{c}{\textit{PSNR}} &  &  &  & \multicolumn{1}{c}{\textit{SSIM}} &  &  &  & \multicolumn{1}{c}{\textit{R}} \\ \midrule
\textit{CUHK (6)} &  &  &  & \textbf{14.6213 $\pm$ 0.4061} &  &  &  & 0.5358 $\pm$ 0.0216 &  &  &  & \textbf{0.8295 $\pm$ 0.0200} \\
\textit{AR (6)} &  &  &  & 14.1721 $\pm$ 0.4127 &  &  &  & \textbf{0.5608 $\pm$ 0.0232} &  &  &  & 0.7811 $\pm$ 0.0217 \\
XM2GTS (6) &  &  &  & 11.7158 $\pm$ 1.3050 &  &  &  & 0.4096 $\pm$ 0.0258 &  &  &  & 0.3817 $\pm$ 0.1341 \\
\textit{All (18)} &  &  &  & 13.5030 $\pm$ 0.5639 &  &  &  & 0.5021 $\pm$ 0.0205 &  &  &  & 0.6641 $\pm$ 0.0658 \\ \bottomrule
\end{tabular}
}
\caption{Comparison of physical (PSNR), perceptual (SSIM) and correlational (R) quality measures for the inverse sketches synthesized from the sketches in the CUFS database and its sub-databases with the line sketch model trained using feature loss alone. $x \pm m$ shows the mean $\pm$ the bootstrap estimate of the standard error of the mean.}
\label{table_5}
\end{table}

\end{document}